\documentclass{article}

\usepackage[utf8]{inputenc} 
\usepackage[T1]{fontenc}    
\usepackage{hyperref}       
\usepackage{url}            
\usepackage{booktabs}       
\usepackage{amsfonts}       
\usepackage{nicefrac}       
\usepackage{microtype}      
\usepackage{cuted}

\usepackage{natbib}

\usepackage[accepted]{aistats2020_preprint}

\usepackage{amsmath}	    
\usepackage{graphicx}
\usepackage{amsthm}
\usepackage{mathtools}
\usepackage{algorithm}
\usepackage{algpseudocode}
\usepackage{wrapfig}
\usepackage{thmtools, thm-restate}
\usepackage{enumitem}

\usepackage{multicol}

\newcommand{\argmin}{\text{argmin}}
\newcommand{\argmax}{\text{argmax}}
\newcommand{\cone}{\text{cone}}
\newcommand{\conv}{\text{conv}}

\newtheorem{innercustomthm}{Theorem}
\newenvironment{customthm}[1]
  {\renewcommand\theinnercustomthm{#1}\innercustomthm}
  {\endinnercustomthm}
  
\DeclareMathOperator*{\trace}{trace}

\newtheorem{definition}{Definition}
\newtheorem{assumption}{Assumption}
\newtheorem{theorem}{Theorem}
\newtheorem{corollary}{Corollary}

\begin{document}
\twocolumn[
\aistatstitle{Rarely-switching linear bandits: optimization of causal effects for the real world}
\aistatsauthor{Benjamin J. Lansdell}
\aistatsaddress{University of Pennsylvania\\USA}
\aistatsauthor{Sofia Triantafillou}
\aistatsaddress{University of Pittsburgh\\USA}
\aistatsauthor{Konrad P. Kording}
\aistatsaddress{University of Pennsylvania\\USA}]

\begin{abstract}
Excessively changing policies in many real world scenarios is difficult, unethical, or expensive. After all, doctor guidelines, tax codes, and price lists can only be reprinted so often. We may thus want to only change a policy when it is probable that the change is beneficial. In cases that a policy is a threshold on contextual variables we can estimate treatment effects for populations lying at the threshold. This allows for a schedule of incremental policy updates that let us optimize a policy while making few detrimental changes. Using this idea, and the theory of linear contextual bandits, we present a conservative policy updating procedure which updates a deterministic policy only when justified. We extend the theory of linear bandits to this rarely-switching case, proving that such procedures share the same regret, up to constant scaling, as the common LinUCB algorithm. However the algorithm makes far fewer changes to its policy and, of those changes, fewer are detrimental. We provide simulations and an analysis of an infant health well-being causal inference dataset, showing the algorithm efficiently learns a good policy with few changes. Our approach allows efficiently solving problems where changes are to be avoided, with potential applications in medicine, economics and beyond.
\end{abstract}

\section{Introduction}

\begin{figure*}[t]
\centering
\includegraphics[width=0.9\textwidth]{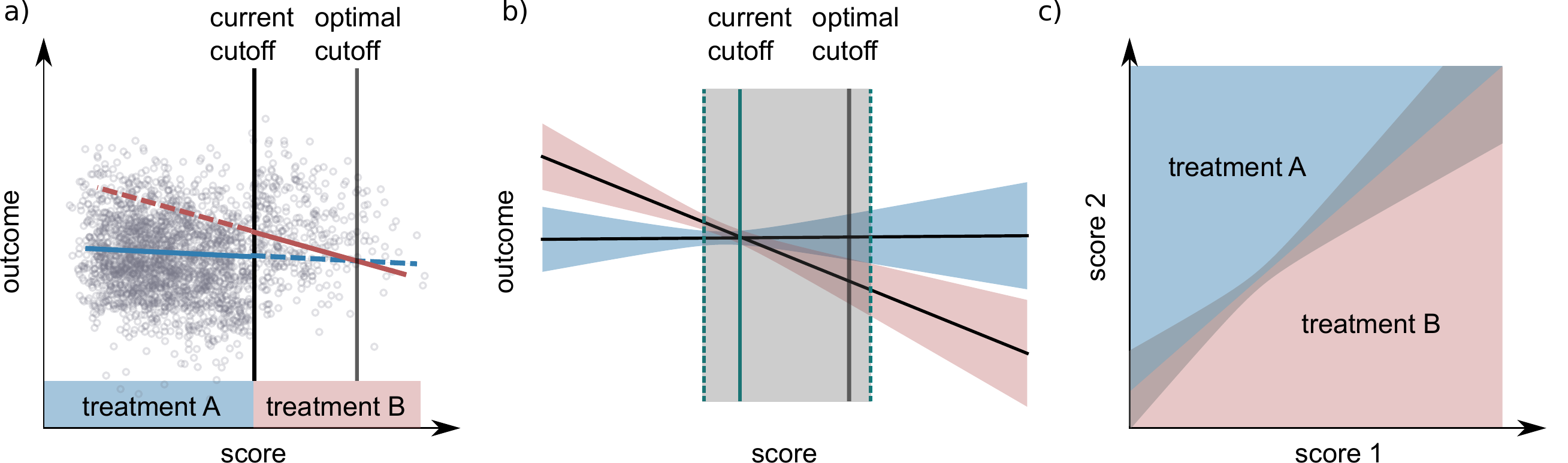}
\caption{a) Many policies assign a treatment by thresholding on a single covariate \citep{Moss2014}. b) A threshold policy may be parameterized by the maximum expected outcome according to an estimated linear model (current line, current cutoff). By construction, with high probability the optimal decision boundary (black vertical line) lies somewhere in a region determined by the confidence intervals on the outcome of each treatment (grey shaded area). For a given context, certain arms cannot plausibly be optimal. An algorithm that never plays these excluded arms is \emph{feasible}. In higher dimensions, policies may assign treatment by thresholding on a linear combination of multiple covariates c) \citep{Bertanha2019, Cattaneo2019}, or assign multiple treatments by thresholding on multiple linear combinations of covariates \citep{Papay2011}. As in the one dimensional case, with high probability the optimal decision boundary lies in a region determined by confidence intervals on the linear parameters estimated for each arm (grey shaded region).}
\label{fig:schematic}
\end{figure*}

Many decisions in healthcare, economics and beyond are made by thresholding on a linear combination of contextual variables \citep{Moscoe2015,Bor2014}: scholarships are awarded when exam scores exceed a threshold \citep{Thistlewaite1960}, assistance is given to those below an income threshold, and many medical treatments and diagnoses are based on thresholding biometrics \citep{Nayor2017, Dennison-himmelfarb2014,Barry} (Figure \ref{fig:schematic}a). Such policies are often applied where it is difficult or unethical to experiment with the population and a policy should benefit the population as much as possible. Exploration is avoided: if a patient arrives with high blood pressure then they receive treatment -- there is no room to explore, hoping to learn more about the treatment's effect. Prior studies will typically have established treatment benefits. These policies, often expressible as simple thresholds, are used when only the population above a threshold is expected to benefit from the treatment. While these policies are generally chosen based on domain knowledge, they could be further optimized (Figure \ref{fig:schematic}b).

Optimizing such policies while acting greedily is challenging. Particularly because, in cases where policies are health or economic recommendations, it may be difficult or costly to change a policy excessively. There is an added need to find an optimal policy in a way that minimizes both the number of erroneous changes and the number of total changes to the policy. In these cases changes to a policy should only be made when there is sufficient evidence it will improve outcomes -- they should be \emph{rarely switching} \citep{Abbasi-yadkori2011}. There is thus a conflict between the greedy aims of such a policy and the need to explore to optimize it.

The balance between exploration and exploitation is well studied in multi-arm bandits. In a bandit problem an agent must choose from a set of actions at each round and only observes reward for the action chosen \citep{Szepesvari2018}. A contextual bandit provides the agent with side information about the set of choices at each round, which can be used to make a more informed decision \citep{Zhou2015a}. Contextual multi-armed bandits are used in personalized medicine, targeted advertising and website design. Modelling the relationship between context and outcome allows for the extrapolation of outcomes in novel contexts and thus, if sufficiently confident about the extrapolation, can obviate the need to explicitly explore.

For instance, it is well known from econometrics that local treatment effects can be estimated from policies that implement thresholds, known as regression discontinuity design (RDD) \citep{Thistlewaite1960}, and by extrapolation these estimated effects can suggest changes to a policy that are highly likely to improve outcomes \citep{Dong2015a}. We propose a more general version of the same idea: maintain a fixed policy until sufficiently confident, on the basis of a contextual model, that a different one would be better. We can draw from multi-armed bandit theory in order to implement and understand the performance of this simple idea.

As machine learning and AI are increasingly used to aid decision making, there is a growing need for safety and performance guarantees to be understood and implemented \citep{Berkenkamp2015,Aboutalebi2019}. Thus a lot of prior work considers conservative or safe bandit algorithms \citep{Wu2016,Katariya2018,Kazerouni2017,Kakade,Sun2016a,Chow2019,Vakili2016, Sani2013, David2016, Zimin2014}. In particular, recent work has considered exploration-free, or greedy bandit algorithms, in a linear contextual setting \citep{Bastani2017,Kannan2018}. Surprisingly, under certain distributional assumptions, noise alone provides the necessary exploration to allow learning for free \citep{Bastani2017}. In fact, a recent comparison between contextual bandit algorithms shows that greedy algorithms can do quite well in practice \citep{Bietti2018, Foster2018} -- exploration bonuses can indeed be avoided in some settings. As the motivating examples involving thresholds introduced above can be thought of as fixed policies that choose based on a linear combination of contextual variables, here we investigate rarely-switching policy optimization in a linear contextual bandit setting. Under the assumption of linearity, on what schedule should a fixed policy be updated to incur low regret?

Only limited previous work has considered rarely-switching policies. For instance, the case where there is a cost to changing actions has received some attention \citep{Guha2009,Koren2017,Koren2017a}. While some of these results could possibly be used here -- by considering a formulation where the space of possible policies is considered the space of arms, and a cost for switching policy becomes the same as a cost for switching action -- doing so would not yield a linear bandit, making the regret behavior more difficult to characterize. Abbasi-Yadkori et al 2011 \citep{Abbasi-yadkori2011} consider a rarely-switching class of linear upper confidence bound (LinUCB) algorithms. However their aim is to save computation, and not to minimize the number of changes made to avoid exploration with policy. Thus, rather than considering cases where there is an explicit cost to change actions, we consider the case where the aim is simply to only change policy when given good evidence it will lead to improvement.

Here we study how to find optimal policies in a linear contextual bandit setting, with a family of rarely-switching policies defined by a fixed linear function of covariates. We analyze these algorithms by extending the theory of the linear upper confidence bound (LinUCB) algorithm. The algorithms, up to constants, share its asymptotic regret behavior \citep{Abbasi-yadkori2011}. To prove that some updates to our policies improve regret with high probability we introduce a novel correspondence with linear programming. Empirically, we show that these methods compare favorably to previous a linear bandit algorithm that rarely switches its policy \citep{Abbasi-yadkori2011}, and to a conservative algorithm that aims to always perform better than a baseline policy \citep{Kazerouni2017}.

\subsection{Linear contextual bandits}
\label{sec:bandits}

We study the following version of the common stochastic linear bandit \citep{Szepesvari2018,Abbasi-yadkori2011}. At each round, $t$, the agent observes a context $\mathbf{s}_t\in\mathcal{D}\subseteq \mathbb{R}^d$ and then selects an action $a_t\in\mathcal{A}$. We assume each context is drawn independently from a distribution $\rho$. The context may represent a feature vector from an underlying state: $\mathbf{s}_t = \phi(\mathbf{x}_t)$. Here we consider cases where each arm is presented with the same information (e.g. when $\mathbf{s}_t$ represents patient information and arms are a fixed set of treatment options -- similar to the set up of \citep{Bastani2017}). This can be considered a special case of the framework in which the function $\phi$, and hence the context $\mathbf{s}_t$, also depend on action $a_t$ \citep{Szepesvari2018}. The agent observes a reward linearly dependent on the features
$$
y_t = \mathbf{s}_t^\mathsf{T}\theta^{a_t} + \eta_t,
$$
for $\theta^{a_t}\in\mathbb{R}^d$ the unknown reward parameters. Let $\theta = (\theta^a)_{a\in\mathcal{A}}$ be the set of all parameters. Let $z_t = \mathbf{s}_t^\mathsf{T} \theta^{a_t}$ denote the expected reward, and let $\mathcal{F}_t$ be the filtration capturing the history of the process up until observing reward at round $t$: $\mathcal{F}_t = \sigma(a_{1}, \dots, a_{t}, \eta_{1}, \dots, \eta_{t-1})$.

We make the following standard assumptions:
\begin{assumption}
The noise $\eta_t$ is conditionally $\sigma$-subgaussian:
$$
\mathbb{E}\left(e^{c\eta_t}|\mathcal{F}_t\right) \le \exp(c\sigma^2/2), \quad \forall c \in \mathbb{R}.
$$
\end{assumption}
The problem is bounded in the following way:
\begin{assumption}
There exist constants $B,D \ge 0$ such that $\|\theta^a\|\le B$, $\|\mathbf{s}_t\|\le D$ and $\|\mathbf{s}_t^\mathsf{T}\theta^{a}\|\in [0,1]$ for all $t$ and $a\in\mathcal{A}.$
\end{assumption}

At each round, for the observed context, there is an optimal arm to pull: $a^*_t = \argmax_{a}\mathbf{s}_t^\mathsf{T} \theta^{a}$, splitting ties arbitrarily. From this we can define the instantaneous regret:
$$
r_t = \mathbf{s}_t^\mathsf{T} (\theta^{a^*_t} - \theta^{a_t}).
$$

We aim to find a policy $\pi:\mathcal{D}\to \mathcal{A}$ that minimizes the cumulative regret over $T$ rounds:
$$
\hat{R}_T = \sum_{t=1}^T r_t.
$$
Let $R_T = \mathbb{E}(\hat{R}_T)$, where the expectation is taken over histories up to horizon $T$.

\subsection{Rarely-switching policies}

The optimal policy is a multiclass classifier:
\begin{equation}
\label{eq:optpol}
\pi^*(\mathbf{s}_t) = \argmax_a \mathbf{s}_t^\mathsf{T}\theta^{a}.
\end{equation}
Here we consider learning policies whose basic form is the following:
\begin{equation}
\label{eq:policy}
\pi(\mathbf{s}_t) = \argmax_a \mathbf{s}_t^\mathsf{T}\tilde{\theta}_t^{a},
\end{equation}
and thus that are parameterized by the set $\tilde{\theta}_t = \{\tilde{\theta}^a_t\}_{a\in\mathcal{A}}$, which can be held constant between rounds. We call this family of policies \emph{rarely-switching} policies. They can be seen as policies in which decisions are made by thresholding on a linear combination of context variables, which is only updated rarely.

\subsection{Confidence bounds}

Rarely-switching policies operate based on the provisionally optimal arm given a set of operational parameter values $\tilde{\theta}$. A policy needs to update these parameters on some schedule to be close to current estimate of the true parameters. We thus maintain the least squares estimate of parameters given the pulls of each particular arm, $\hat{\theta}^a_t$, defined below. Let $T_t^a$ be the set of times for which arm $a$ was pulled up to and including round $t$. Then let the arm parameter estimates be given by the ridge regression solution
\begin{align*}
V_t^a &= \lambda I + \sum_{s\in T_t^a} \mathbf{s}_s\mathbf{s}_s^\mathsf{T},\\
\hat{\theta}_t^a &= (V_t^a)^{-1}\sum_{s\in T_t^a}y_s\mathbf{s}_s,
\end{align*}
for regularization parameter $\lambda > 0$. Let $V_t$ be the block diagonal matrix constructed from the matrices $\{V_t^a\}$. To construct working rarely-switching bandit algorithms we also must construct confidence bounds for each arm. Care is needed in constructing confidence sets for these estimators because, unlike in standard least squares estimation, the sequence of observations $\{\mathbf{s}_s, y_s\}_{s=1}^t$ are no longer independent and identically distributed -- the action chosen depends on the history of the process, which induces correlations among $\{\mathbf{s_t}\}_{t\in T_t^a}$. By using Martingale theory, we can find confidence intervals that apply for any policy \citep{Szepesvari2018}. Let $\mathcal{C}_t$ be the following set:
$$
\mathcal{C}_t = \{\mathbf{x}\in\mathbb{R}^d:\|\hat{\theta}^a_{t-1} - \mathbf{x} \|^2_{V_{t-1}} \le \beta_t\},
$$
where $\|\mathbf{x}\|^2_V = \mathbf{x}^\mathsf{T}V\mathbf{x}$, for positive definite matrix $V$, and some bound $\beta_t$. We can place a lower bound on the probability that the true parameters lie in the set of confidence sets for all rounds. Let this be the event:
$$
\mathcal{E}_t = \cap_{n = 1}^t\{\theta \in \mathcal{C}_n\}.
$$
Theorem 20.2 of \citep{Szepesvari2018} provides an explicit form of $\beta_t$ for which the event $\mathcal{E}_t$ occurs with probability at least $1-\delta$ (provided as Theorem B1 in the supplementary material).

\subsection{Feasible contextual linear bandits}

Which rarely-switching algorithms learn the optimal parameters? We answer this question by first studying a more general class of algorithms. Consider the general case of the $k$-armed contextual bandit, $\mathcal{A} = \{1,\dots,k\}$. For each observed context, $\mathbf{s}_t$, the confidence bounds define plausible intervals of payoffs for each arm. That is, assuming $\mathcal{E}_t$ occurs, the true expected reward for arm $a$ lies in the interval
$$
\mathbf{s}^\mathsf{T}\hat{\theta}^a_t \pm \sqrt{\beta_t}\|\mathbf{s}\|_{(V_t^a)^{-1}}.
$$
For each round, assuming the confidence bounds hold, the upper bound payoff of some arms may be below the lower bound payoff of another arm, excluding these arms from plausibly being the best choice (Figure \ref{fig:schematic}b,c). This motivates to the following definition:
\begin{definition}
For a context $\mathbf{s}_t$, there is a set of arms which are plausibly the optimal arm $\mathcal{J}(\mathbf{s}_t) \subseteq \mathcal{A}$. This is the set of valid arms that are not excluded from being optimal due to their confidence bounds. An algorithm that only plays arms in $\mathcal{J}(\mathbf{s}_t)$, for all rounds $t$, is called a \emph{feasible algorithm}. \end{definition}
With this we can establish the following result:
\begin{theorem}
With probability at least $1-\delta$, the regret of any policy learned by a feasible algorithm is bounded by
$$
\hat{R}_T \le \sqrt{32dT\beta_T\log\left(\frac{\trace(V_0) + TL^2}{d\det^{1/d}(V_0)}\right)}.
$$
\end{theorem}
All proofs are provided as supplementary material. Corollary 19.3 from \citep{Szepesvari2018} can be applied and then provides:
\begin{corollary}
Choosing $\delta = 1/T$, the expected regret obeys
$$
R_T \le Kd\sqrt{T}\log(TL),
$$
for constant $K > 0$.
\end{corollary}

Note that the LinUCB algorithm and the greedy algorithm ($\tilde{\theta}_t = \hat{\theta}_t$, see, for example, Bastani et al \citep{Bastani2017}) are feasible algorithms.

\section{Feasible, rarely-switching bandits}

Given this sublinear regret behavior, we will thus construct feasible, rarely-switching algorithms. To derive them we first consider the geometry of the linear contextual bandit.

\begin{figure}[t]
\centering
\includegraphics[width=.45\textwidth]{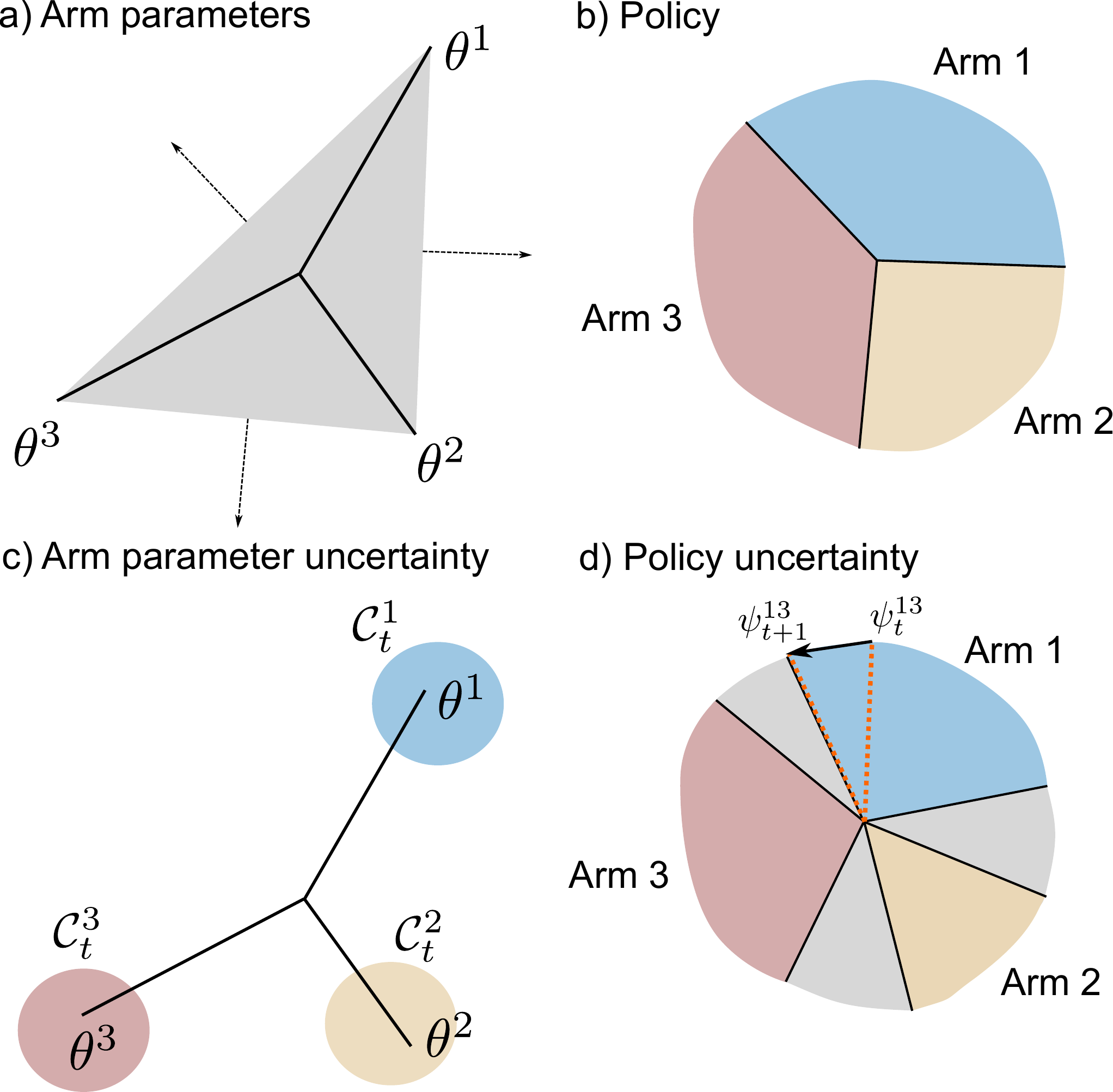}
\caption{The geometry of linear contextual bandits. (a) The arm parameters define a convex polytope which determines the policy (b). In this example, the decision boundaries are normal vectors to the faces of $C$. (c) Uncertainty in the parameters creates uncertainty in the polytope, (d) which in turn creates uncertainty in the policy (grey areas). When a decision boundary is definitely outside the plausible interval, moving it to the edge of that interval will improve expected regret of the policy.}
\label{fig:convexcones}
\vspace{-10pt}
\end{figure}

For each context, $\mathbf{s}_t$, policies of the form \eqref{eq:optpol} are linear programs \citep{murty1983linear}. That is, they are solutions to:
\begin{equation}
\label{eq:linprog}
\max_{\mathbf{x}\in\conv(\theta)} \mathbf{s}_t^\mathsf{T} \mathbf{x},
\end{equation}
where $\conv(\theta)$ denotes the convex hull of the set of arm parameters $\{\theta^a\}_{a=1}^k$, and can equivalently be represented as a set of linear inequalities $A\mathbf{x} \le \mathbf{b}$. $\conv(\theta)$ is a closed convex polytope in $\mathbb{R}^d$. For all $\mathbf{s}_t\in\mathbb{R}^d \setminus \{0\}$, a solution to \eqref{eq:linprog} can found on one of the vertices of $\conv(\theta)$ (\citep{murty1983linear}, Theorem 3.3). Thus a solution to \eqref{eq:optpol} is a solution to \eqref{eq:linprog} and at least one solution of \eqref{eq:linprog} is a solution to \eqref{eq:optpol}. This correspondence provides a lot of structure. For instance, it is clear that arms which are not extreme points of $\conv(\theta)$ are never played. And, the decision boundaries of $\pi(\mathbf{s}_t)$ can be seen as contexts $\mathbf{s}_t$ which define supporting hyperplanes which contain more than one extreme point (Figure \ref{fig:convexcones}a,b). The uncertainty in parameter estimates $\hat{\theta}$ induces uncertainty in the best policy (Figure \ref{fig:convexcones}c). Such problems are studied in robust optimization \citep{Ben-Tal2009,Bertsimas2004}. 

Importantly, note that an algorithm is feasible if it never plays an excluded arm. The confidence sets are constructed such that the decision boundaries lie in \emph{decision regions} (Figure \ref{fig:convexcones}d) with high probability. For two arms, $i$ and $j$, let
$\tilde{\theta}_t^{i}-\tilde{\theta}_t^{j}\coloneqq \psi_t^{ij}$ and let $\Psi^{ij}_t$ be the set difference $\Psi^{ij}_t = \{\psi\in\mathbb{R}^d| \psi = \theta^i - \theta^j, \theta \in \mathcal{C}_t\}$. With this, we can prove:
\begin{theorem}
Any rarely-switching algorithm that maintains $(\tilde{\theta}_t^{i}-\tilde{\theta}_t^{j})\in\cone(\Psi_t^{ij})$ for all $i,j\in[k]$ and for all $t$ is a feasible algorithm.
\end{theorem}
This means that algorithms which maintain plausible decision boundaries are feasible algorithms, and share the above sub-linear regret bound.

\subsection{The make-no-mistakes maxim}

We seek rarely-switching policies that only change when justified -- when their behavior is plausibly sub-optimal. The above considerations suggest that the policy need not be changed when $\psi_t^{ij}\in\cone(\Psi_t^{ij}), \forall i,j\in[k]$. When this is not the case, how should the policy be updated? There are a few options based on the exact sense in which the policy should be conservative. We discuss two such ideas. One idea is that updates to the policy should be as small as possible to maintain feasibility, thus minimizing the chance that the update will in fact lead to higher expected regret. Such a policy seeks to minimize mistakes when changing policy. Since the magnitude of $\psi_t^{ij}$ does not matter, we can consider the smallest update to the policy to be the update that keeps $\tilde{\theta}_t\in\mathcal{C}_t$ and that minimizes the \emph{angle} between $\psi_t^{ij}$ and $\psi_{t+1}^{ij}$.

\subsection{The conservative rarely-switching bandit}

The idea of minimizing the angle between each decision boundary pair suggests updating $\tilde{\theta}_t$ as follows:
$$
\tilde{\theta}_{t+1} = \argmax_{\varphi\in\mathcal{C}_t}\frac{\tilde{\theta}_t^\mathsf{T} F^\mathsf{T}F\varphi}{\|F \tilde{\theta}_t\|\|F\varphi\|} \coloneqq \argmax_{\varphi\in\mathcal{C}_t}K(\varphi),
$$
where $F$ is the block matrix that computes the decision boundaries for each pair of arms. That is, $F\varphi$ returns $\psi_{t}^{ij}$ for all pairs $i < j$. We call this the \emph{conservative rarely-switching} bandit. Let $J \coloneqq F^\mathsf{T}F$. Then this can be understood as a minimization of the cosine distance under inner product $x^\mathsf{T}Jy$ (see, for example, \citep{Wua,Choi}). It could be solved with geodesic-convex routines \citep{Zhang2016c,Ferreira2014}. Here we solve the problem using a projected gradient-descent method (Algorithm \ref{al:racgp}, more details in supplementary material). 

\begin{algorithm}[t]
\caption{The conservative, rarely-switching bandit}
\label{al:racgp}
\begin{algorithmic}
\Require initial policy $\tilde{\theta}_0$, regularization parameter $\lambda$, number of rounds $T$, initial gradient step size $\epsilon$, tolerance $\Delta$, number of projected gradient iterations $n_{iter}$.
\State {Initialize $V_0 \gets \lambda I$ \;}
\For{$t\in [0, T]$}
    \State {Observe context: $\mathbf{s}_t$ \;}
    \State {Choose action: $a_t \gets \argmax_a \mathbf{s}_t^\mathsf{T}\tilde{\theta}_{t-1}$\;}
    \State {Obtain reward $y_t$\;}
    \State {Update parameters: $V_t$, $U_t$, $\hat{\theta}_t$, $\beta_t$ and bounds $\mathcal{C}_t$\;}
    \State {$\varphi_0 \gets \tilde{\theta}_{t-1}$ \;}
    \For{$i\in [1, n_{iter}]$}
      \State {Set step size: $\eta \gets \epsilon / \sqrt{i}$ \;}
      \State {Gradient step, using Suppl. Eq. (2): $\varphi' \gets \varphi_{i-1} + \eta\frac{\partial K}{\partial \varphi}(\varphi_{i-1})$ \;}
      \State {Project onto $\mathcal{C}_t$: $\varphi_i \gets \argmin_{\phi \in\mathcal{C}_t} \|\phi - \varphi'\|_{V_t}^2$ \;}
    \EndFor
    \State {boundary\_angles $\gets \frac{\tilde{\theta}_{t-1}^\mathsf{T} J \varphi_{n_{iter}}}{\|\tilde{\theta}_{t-1}\|_J \|\varphi_{n_{iter}}\|_J}$\; }
    \If{boundary\_angles $ < 1 - \Delta$ and $\|\tilde{\theta}_{t-1} - \hat{\theta}_t\|_{V_t} > \beta_t$}
    \State {$\tilde{\theta}_{t} \gets \varphi_{n_{iter}}$\;}
    \Else
    \State {$\tilde{\theta}_{t} \gets \tilde{\theta}_{t-1}$\;}
    \EndIf
\EndFor
\end{algorithmic}
\end{algorithm}

\subsection{The greedy rarely-switching bandit}

An issue with the conservative update is that, by aiming to place the decision boundaries on the edge of what is feasible, they may need to be updated again in the next few rounds -- contrary to the aim of having a policy that does not need to be updated excessively. An alternative is to update the parameter with greedy updates: whenever $T\tilde{\theta}_t\notin \cone(T\Psi_t)$, then set $\tilde{\theta}_t = \hat{\theta}_t$. This modifies the algorithm in an obvious way (see supplementary material). We call this the \emph{greedy rarely-switching} bandit. It sacrifices confidence that a given update will definitely be an improvement for an algorithm that makes fewer updates to the policy.

\subsection{Simpler rarely-switching feasible algorithms}

As a comparison, we also consider the simpler algorithm which involves changing policy whenever the parameter estimates move out of the confidence intervals, as opposed to the above algorithms, which change policy whenever the \emph{decision boundaries} are no longer feasible. The conservative version of this idea sets:
$$
\tilde{\theta}_t = \argmin_{\varphi \in \mathcal{C}_t} \|\varphi - \tilde{\theta}_{t-1}\|_{V_t}.
$$
That is, as soon as the parameters $\tilde{\theta}_t$ are not feasible then the algorithm updates $\tilde{\theta}$. The greedy version of this idea is:
$$
\tilde{\theta}_t = 
\begin{cases}
\tilde{\theta}_{t-1},& \tilde{\theta}_{t-1} \in\mathcal{C}_t\\
\hat{\theta}_t, \quad \text{else}
\end{cases}.
$$
To distinguish these algorithms from the rarely-switching versions, we call these the feasible conservative, and feasible greedy algorithms, respectively.\footnote{We perform one additional sanity check by comparing these algorithms to a deterministically switching algorithm that updates its policy on a fixed schedule, presented in the supplementary material. The comparison shows it is indeed worthwhile considering adaptive rarely-switching algorithms.}

\subsection{High probability improvement in two dimensions}

Here we prove that, in the planar case ($d=2$), the idea of minimizing the change in angle of the decision boundaries can lead, in identifiable circumstances, to a decrease in the expected regret at each round (with probability at least $1-\delta$). The idea is that when decision boundaries are outside a plausible interval, moving the boundary to the edge of this interval only affects the behavior of the policy in a region where the arm to pull is clear -- and thus the policy moves from definitely incurring regret to not necessarily incurring regret for contexts in this region (Figure \ref{fig:convexcones}d).

In order to prove these updates never increase regret, we note that, in two dimensions, the policy $\tilde{\theta}_t$ defines intervals over angles in which each arm is played. The policy thus becomes one dimensional -- it can be considered as a cyclic ordering of the arms based on angle, $\sigma = (\sigma(\tilde{\theta}_1)\dots\sigma(\tilde{\theta}_k))$, and a set of angles corresponding to the decision boundaries between the pair of arms $\tilde{\theta}^{\sigma_{i}}$ and $\tilde{\theta}^{\sigma_{i+1}}$. If, for a given change in policy, $\tilde{\theta}_t\to\tilde{\theta}_{t+1}$ we consider also the set of angles, $\Omega$, that describe decision boundaries in both the old and the new policies, then we have:
\begin{theorem}
For $d = 2$, if an update to the policy $\tilde{\theta}_t$ satisfies:
\begin{enumerate}[label=(\roman*)]
    \item the set of playable arms stays the same between $t$ and $t+1$ -- that is, an arm does not become dominated (stops being an extreme point of the polytope), or stops being dominated;
    \item the ordering of arms, $\sigma$, in the new and old policies is the same;
    \item for each decision boundary $b = \psi_t^{\sigma_i,\sigma_{i+1}}$, either the element before or after $b$ in $\Omega$ is $\psi_{t+1}^{\sigma_i,\sigma_{i+1}}$;
    \item each change $\psi_{t}^{ij}\to\psi_{t+1}^{ij}$ is the smallest such that $\psi_{t+1}^{ij}\in\cone(\Psi_{t+1}^{ij})$. I.e. that
    $$
    \psi_{t+1}^{ij} = \argmin_{\psi\in\cone(\Psi_{t+1}^{ij})}\text{arccosdist}(\psi,\psi_t^{ij});
    $$
\end{enumerate}
then it does not increase the expected instantaneous regret, for a fixed policy, in the sense that:
$$
\mathbb{E}_{\mathbf{s}\sim \rho}\left(\mathbf{s}^\mathsf{T} \theta^{a_t}\right)
\le \mathbb{E}_{\mathbf{s}\sim \rho}\left(\mathbf{s}^\mathsf{T} \theta^{a_{t+1}}\right),
$$
with probability at least $1-\delta$.
\end{theorem}
Thus, in the common two-armed case, in some cases it may be possible to make updates that maintain feasibility, and that improve expected regret with high probability.

\section{Results}

\subsection{Simulations}

\begin{figure*}[t]
\centering
\includegraphics[width=\textwidth]{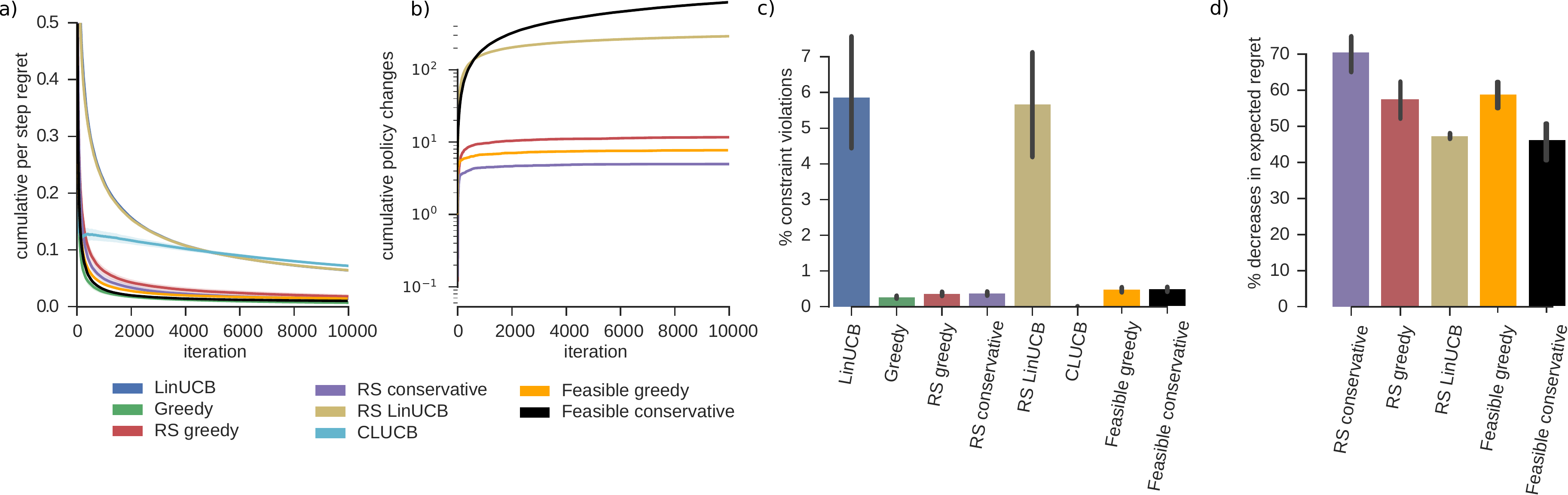}
\caption{Simulation results. Fifty random generated bandit parameters are run for 10000 rounds. a) The per-step regret ($R_t/t$). b) The number of policy changes. Both a) and b) traces show mean, plus/minus standard error. c) The percentage of rounds in which each method performs below the cumulative reward obtained when only a baseline arm is played. d) Percentage changes to policy that decrease expected regret. Error bars indicate 95\% confidence intervals.}
\label{fig:simulated comparisons}
\end{figure*}

To test the conservative and greedy rarely-switching bandit algorithms we generate synthetic data from a set of randomly generated arms. Each arm parameter is sampled randomly from $\mathbb{R}^d$. Each algorithm is run for 10,000 rounds. Here we present results for $k=4$ arms and $d=5$ (additional experiments with other choices are provided in the supplementary material). At each round a context is sampled uniformly from $\mathcal{D}$, and reward is given according to the chosen arm's mean plus Gaussian noise with standard deviation 0.1. We take $\delta = 0.0001$. The bandit algorithms are compared to the LinUCB algorithm \citep{Abbasi-yadkori2011}, the rarely-switching variant of the LinUCB algorithm \citep{Abbasi-yadkori2011}, the greedy least squares algorithm \citep{Bastani2017}, and the conservative LinUCB algorithm (CLUCB) \citep{Kazerouni2017}.

Both rarely-switching (RS) algorithms perform well. First, the per-step regret $(\frac{R_t}{t})$ is lowest for the greedy algorithm, consistent with previous bandit comparisons \citep{Bietti2018} (Figure 3a). However the RS algorithms perform almost as well, followed by the LinUCB and RS LinUCB methods. The CLUCB algorithm is the slowest to learn: despite the fact that its baseline arm is set to the be the arm with the highest expected return, meaning it starts with an already good policy, ultimately the CLUCB method has the highest per step regret. Second, we see that the mean number of changes for both rarely-switching algorithms is very low (Figure 3b). Indeed, the RS-greedy algorithm achieves low regret with as few as five average changes in policy. Both RS policies changes much less than the RS LinUCB algorithm ($\approx 250$). We also compare each algorithm's performance relative to a policy that always chooses a fixed, `baseline' arm (set to be the with the highest expected reward). By rarely updating their policy, the rarely-switching algorithms seldom have cumulative reward below the baseline policy (Figure 3c). In contrast, the LinUCB algorithms perform below the baseline rate on average around 6\% of the time, while the CLUCB algorithm, by its design, essentially never performs below the baseline. Finally, we examine the proportion of changes which result in lower expected regret for each method (Figure 3d). Consistent with our reasoning, a higher proportion of policy changes made by the RS-conservative algorithm do indeed lead to improved expected regret, compared with changes made by either RS-greedy or RS-LinUCB (both around chance). Taken together, these result show that rarely updating a bandit policy can result in efficient and safe learning, with only a very small number of changes to the policy.

We also compare the rarely-switching algorithms to the feasible algorithms. The feasible algorithms have good per-step regret on the same simulated tasks (Figure \ref{fig:simulated comparisons}a). However, the number of changes to the policy is greater than the rarely-switching conservative and greedy versions (Figure \ref{fig:simulated comparisons}b). Further, the feasible-conservative algorithm does not produce updates which have a good chance of leading to decreased regret, in contrast to the rarely-switching conservative algorithm (Figure \ref{fig:simulated comparisons}d). Thus considering only updating policy when the decision boundary becomes infeasible, as opposed to the arm parameters, is worthwhile. 

\begin{figure*}[t!]
\centering
\includegraphics[width=.65\textwidth]{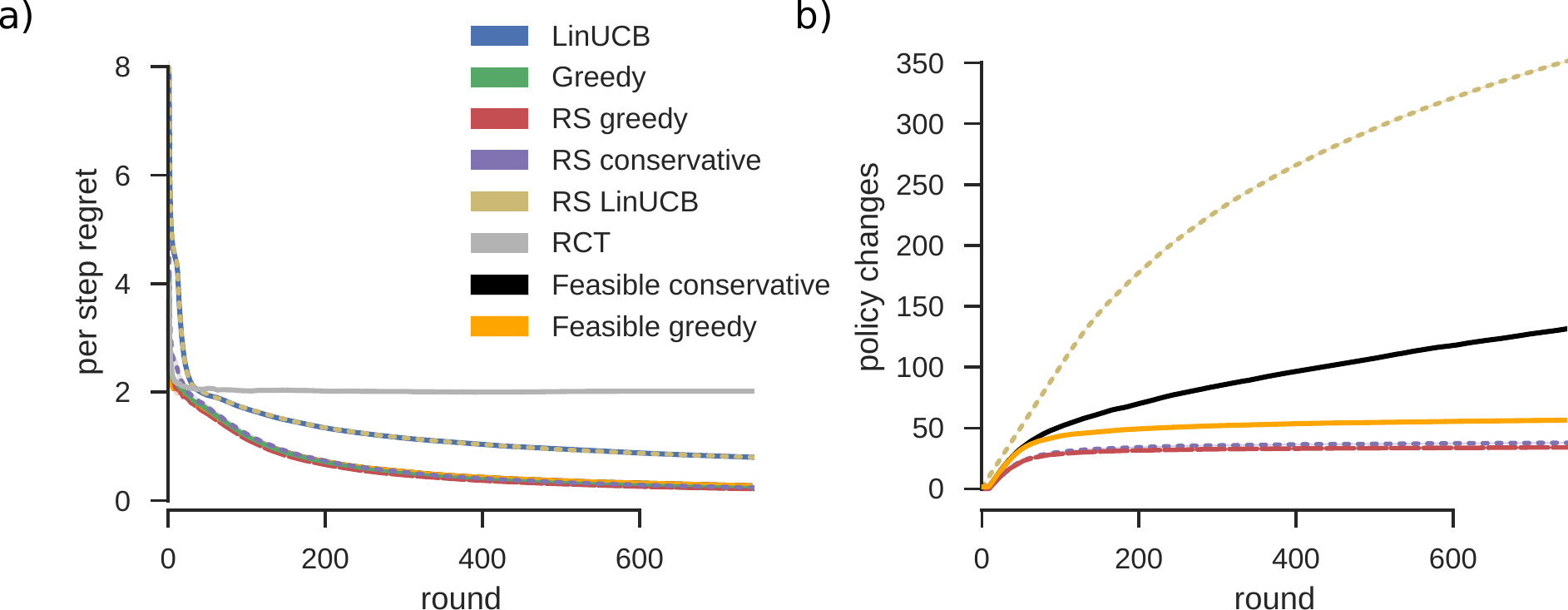}
\caption{Data results. a) Per-step regret for each method over 100 realizations of IHDP semi-simulated trial. The per-step regret for a totally random policy (an RCT) is shown for comparison. b) Number of cumulative policy changes for each method that rarely switches policy. Dashed lines are plotted for some methods for clarity.}
\label{fig:ihdp}
\end{figure*}
\subsection{Real data}
Next we apply the algorithms to a medical dataset: a study on the effect of high-quality child care and home visits on future cognitive performance. By presenting subjects sequentially to each algorithm, we test if the algorithms can be used to allocate treatment to subjects such that as many benefit as possible -- or least more than would benefit from a randomized control trial (RCT). We use a semi-simulated infant health and development program (IDHP) dataset, in which counterfactuals are simulated from a learned model from the actual trial \citep{Hill2011b}. This dataset has been used as a benchmark for causal learning (e.g. \citet{Shalit2016a}). The dataset consists of 747 subjects, and 100 simulated outcomes and counterfactuals. We observe that all algorithms learn an optimal treatment within the trial. However the rarely-switching and greedy algorithms learn significantly faster than the LinUCB and rarely-switching LinUCB (Figure 4a). Further, both rarely-switching algorithms learns a good policy with approximately 35 changes in policy (Figure 4b). In comparison, the feasible algorithms have similar regret to the rarely switching versions, but change policy significantly more times (Figure \ref{fig:ihdp}). Thus, again, it is useful to update policies by considering when decision boundaries are feasible, and not just the parameters. These results suggest that these algorithms can be used to efficiently learn good policies with minimal changes, even over a short trial -- making them relevant for adaptive clinical trial designs. 

\section{Discussion}

Here we proposed an approach to optimize policies determined by a linear combination of covariates in a rarely-switching manner. Our contribution are as follows: first, we establish a regret bound for a large class of linear bandit algorithms (so-called feasible algorithms). Second, we present two rarely-switching algorithms, a conservative algorithm based on an angle minimization, and a greedy algorithm. Both of these are feasible algorithms and thus satisfy our derived regret bound. We also provide conditions in which, for a given change in a policy in $d = 2$, regret will not increase with high probability. This is based on a geometric view of the problem that may prove useful in analysis of other rarely-switching linear bandit problems. Both in simulation and real data, the methods learn as efficiently as other state-of-the-art bandit algorithms. The greedy rarely-switching approach requires very few changes to reach a good policy, and the conservative policy makes a high proportion of changes to policy that are improvements (lead to lower expected regret). While the CLUCB algorithm more reliably performs above a baseline policy, it is slow to adopt a higher reward arm, resulting in higher regret. The rarely-switching algorithms do not suffer from this shortcoming. Further, by being greedier than LinUCB, RS methods more quickly figure out better arms to play, and thus in general are more quick to perform above a baseline level. Though we do not perform a comprehensive comparison between other bandit algorithms, these findings are consistent with previous bandit studies \citep{Foster2018}.

There are two notable caveats to the rarely-switching approach we developed. First, the algorithms are not guaranteed to converge with almost certainty. For applications where safety is important, this is an important consideration. However this shortcoming is not unique to our algorithms: the LinUCB algorithm can also incur linear regret on a given run with a non-zero probability. We thus believe establishing conditions for convergence with probability 1 is beyond the scope of the current study. Second, in many applications a linear model may not be reasonable. This could be addressed by considering GLMs, or by considering local regression models near the decision boundaries. The optimization of treatment effects by performing local linear regression and adjusting a threshold accordingly is closely related to the regression discontinuity design used in econometrics \citep{Thistlewaite1960}, and to recent efforts to use RDD to optimize thresholds \citep{Dong2015a}. Expanding our analysis to this locally linear case may provide useful theory for optimizing treatments in the many cases where RDD is or could be used to measure causal effects \citep{Moscoe2015,Marinescu2018}.

We have introduced a novel class of rarely-switching bandit algorithms. Many theoretical properties of our framework need to be explored, but are beyond the scope of this initial study. For instance: what is the dependence of the frequency of policy updates as a function of model parameters (dimensions, number of arms, etc); in simplified cases, what is an optimal update schedule; and how close are the algorithms proposed here to the Pareto frontier when framed as a multi-objective optimization? Answers to these questions are the subject of future work. 

In many real world settings, changing policies is expensive or difficult. As the use of machine learning to find optimal policies increases, considering this inertia is increasingly important. Using the theory of contextual bandits, here we provided one of the first analyses of this problem.

\medskip
\small
\bibliographystyle{plainnat}
\bibliography{Writeups-RDD.bib,Writeups-bandits.bib}

\onecolumn 
\appendix
\section*{Supplementary material for Rarely-switching linear bandits: optimization of causal effects for the real world}

\section{Algorithm details}

\begin{figure*}
\centering
\includegraphics[width=0.8\textwidth]{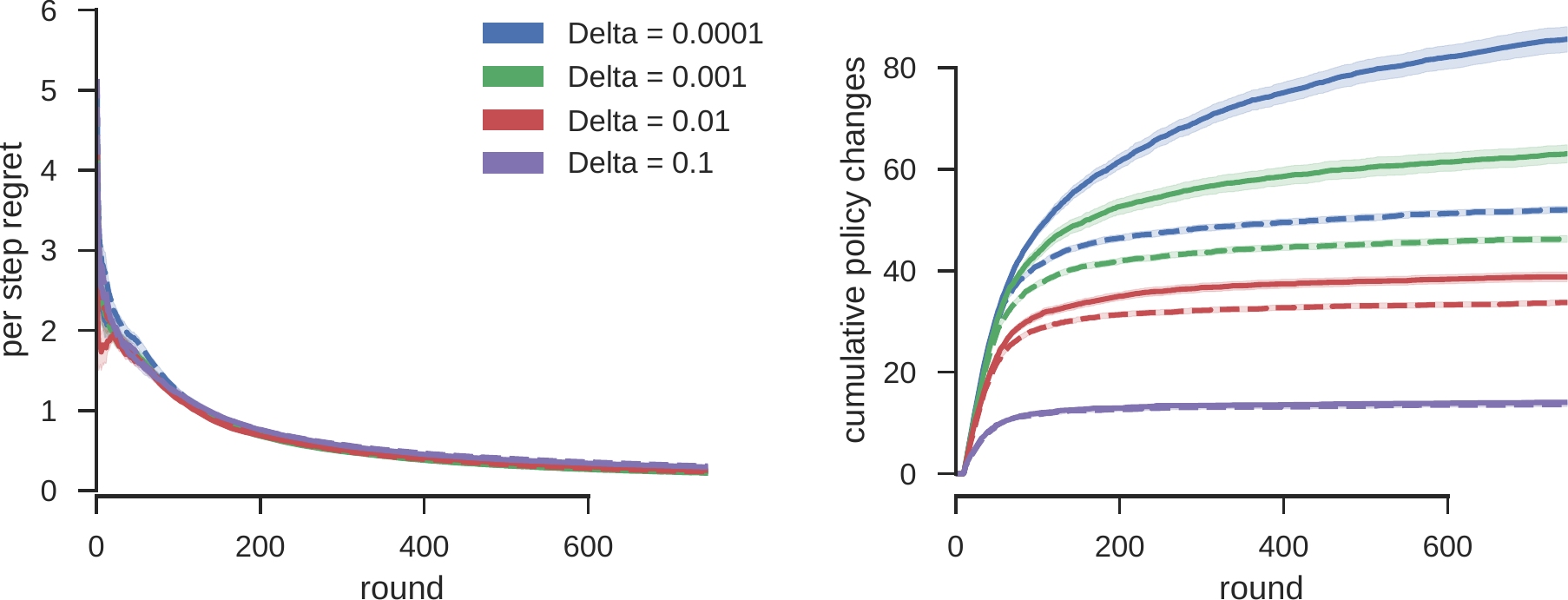}
\caption{Robustness of RS algorithms to value of $\Delta$. On the IHDP dataset, values of delta between $10^{-1}$ and $10^{-4}$ have little impact on the per step regret. But does affect the number of policy changes made. Dashed lines represent RS-greedy, solid lines represent RS-conservative for different $\Delta$ tolerances.} 
\label{fig:del_robust}
\end{figure*}

For the rarely-switching conservative bandit, we implement a projected gradient method. To solve 
$$
\tilde{\theta}_{t+1} = \argmax_{\varphi\in\mathcal{C}_t}\frac{\tilde{\theta}_t^\mathsf{T} F^\mathsf{T}F\varphi}{\|F \tilde{\theta}_t\|\|F\varphi\|},
$$
the method consists of alternating $n_{iter}$ times between gradient updates to maximize:
\begin{equation}
\label{eq:grad}
K(\varphi) = \frac{\tilde{\theta}_t^\mathsf{T} F^\mathsf{T}F\varphi}{\|F \tilde{\theta}_t\|\|F\varphi\|},
\end{equation}
followed by projections onto the convex set $\mathcal{C}_t$. Let $J = F^\mathsf{T}F$. Then the gradient with respect to $\varphi$ is:
\begin{equation}
\label{eq:grad1}
\frac{\partial K}{\partial \varphi}(\varphi) = \frac{1}{\|F \tilde{\theta}_t\|(\varphi^\mathsf{T} J \varphi)^{3/2}}\left(J \varphi (\tilde{\theta}^\mathsf{T} J \varphi) - J\tilde{\theta}(\varphi^\mathsf{T} J \varphi)\right).
\end{equation}
We decide to update the parameters if the smallest angle between $\tilde{\theta}_{t-1}$ and $\varphi\in\mathcal{C}_t$ is above some threshold, $\text{arccos}(\Delta)$, and if the parameters $\tilde{\theta}_{t-1}$ are indeed not in the current confidence set $\mathcal{C}_t$. The algorithm has parameters, $n_{iter}$, step size $\epsilon$, and a tolerance at which we decide two vectors have the same angle, $\Delta$. Here we use $n_{iter} = 100$, $\epsilon = 0.1$ and $\Delta = 0.01$. The algorithm was tested for robustness to the choice in hyperparameters. Particularly $\Delta$ -- values of $\Delta$ between $10^{-1}$ and $10^{-4}$ do not affect the regret behavior (Supplementary Figure \ref{fig:del_robust}).

\begin{figure*}
\centering
\includegraphics[width=0.8\textwidth]{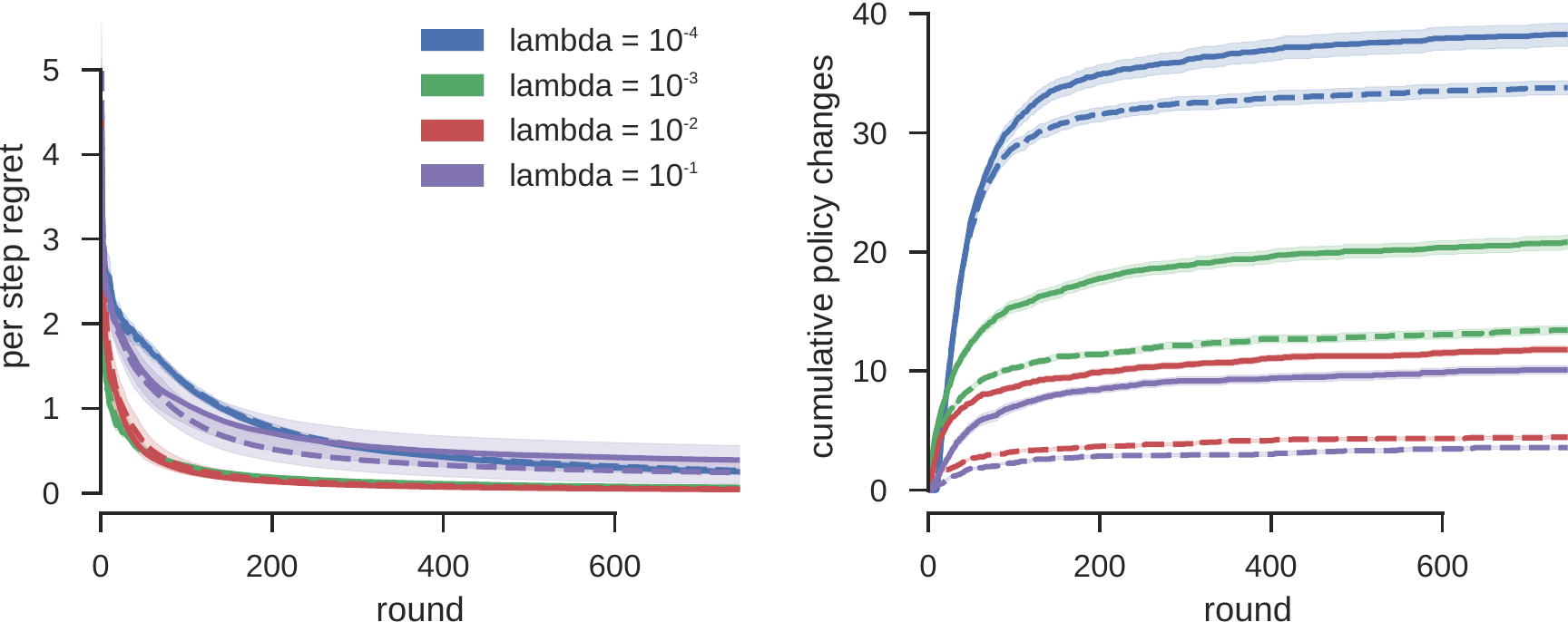}
\caption{Robustness of RS algorithms to value of $\lambda$. On the IHDP dataset, the value of $\lambda$ between $10^{-1}$ and $10^{-4}$ affects both the per step regret and the number of policy changes. Dashed lines represent RS-greedy, solid lines represent RS-conservative for different $\lambda$ values.} 
\label{fig:lmb_robust}
\end{figure*}

The other hyperparameter $\lambda$ determines the amount of regularization in the linear regression. We also checked the algorithm's robustness to varying $\lambda$. In fact there is an optimal $\lambda = 10^{-2}$, in terms of both regret and number of policy changes, performs best or close to best (Supplementary Figure \ref{fig:lmb_robust}).

The rarely-switching greedy algorithm only changes the last section of the algorithm. When an updated is required, it sets $\tilde{\theta}_t = \hat{\theta}_t$ (Supplementary Algorithm \ref{al:racgpgreedy}).

\begin{algorithm}
\caption{The greedy, rarely-switching bandit}
\label{al:racgpgreedy}
\begin{algorithmic}
\Require initial policy $\tilde{\theta}_0$, regularization parameter $\lambda$, number of rounds $T$, initial gradient step size $\epsilon$, tolerance $\Delta$, number of projected gradient iterations $n_{iter}$.
\State {Initialize $V_0 \gets \lambda I$ \;}
\For{$t\in [0, T]$}
    \State {Observe context: $\mathbf{s}_t$ \;}
    \State {Choose action: $a_t \gets \argmax_a \mathbf{s}_t^\mathsf{T}\tilde{\theta}_{t-1}$ and obtain reward $y_t$\;}
    \State {Update parameters: $V_t$, $U_t$, $\hat{\theta}_t$, $\beta_t$ and bounds $\mathcal{C}_t$\;}
    \State {$\varphi_0 \gets \tilde{\theta}_{t-1}$ \;}
    \For{$i\in [1, n_{iter}]$}
      \State {Set step size: $\eta \gets \epsilon / \sqrt{i}$ \;}
      \State {Gradient step, using Equation \eqref{eq:grad}: $\varphi' \gets \varphi_{i-1} + \eta\frac{\partial K}{\partial \varphi}(\varphi_{i-1})$ \;}
      \State {Project onto $\mathcal{C}_t$: $\varphi_i \gets \argmin_{\phi \in\mathcal{C}_t} \|\phi - \varphi'\|_{V_t}^2$ \;}
    \EndFor
    \State {boundary\_angles $\gets \frac{\tilde{\theta}_{t-1}^\mathsf{T} J \varphi_{n_{iter}}}{\|\tilde{\theta}_{t-1}\|_J \|\varphi_{n_{iter}}\|_J}$\; }
    \If{boundary\_angles $ < 1 - \Delta$ and $\|\tilde{\theta}_{t-1} - \hat{\theta}_t\|_{V_t}^2 > \beta_t$}
    \State {$\tilde{\theta}_{t} \gets \hat{\theta}_t$\;}
    \Else
    \State {$\tilde{\theta}_{t} \gets \tilde{\theta}_{t-1}$\;}
    \EndIf
\EndFor
\end{algorithmic}
\end{algorithm}

\section{Proofs of Theorems}

The following is a restatement of Theorem 20.2 from \citep{Szepesvari2018}, and provides an explicit form for the size of the uncertainty sets. We omit the proof here. 

\begin{customthm}{B1}
For any $\delta \in (0,1)$, with probability at least $1-\delta$, it holds that for all $t\in \mathbb{N}_+$,
$$
\|\hat{\theta} - \theta\|_{V_t} < \sqrt{\lambda}\|\theta\| + \sqrt{2\log\left(\frac{1}{\delta}\right) + \log\left(\frac{\det V_t}{\lambda^d}\right)}.
$$
Also, for $\|\theta\| \le L$ then $\mathbb{P}(\mathcal{E}_t) \ge 1-\delta$ with $\mathcal{C}_t$ defined using
$$
\beta_t = \sqrt{\lambda}L + \sqrt{2\log\left(\frac{1}{\delta}\right) + \log\left(\frac{\det V_t}{\lambda^d}\right)}.
$$
\end{customthm}

\begin{customthm}{1}
With probability at least $1-\delta$, the regret of any policy learned by a feasible algorithm is bounded by
$$
\hat{R}_T \le \sqrt{32dT\beta_T\log\left(\frac{\trace(V_0) + TL^2}{d\det^{1/d}(V_0)}\right)}.
$$

\begin{proof}
For a given context $\mathbf{s}_t$ let $\mathcal{J}(\mathbf{s}_t) \subseteq \mathcal{A}$ denote the set of valid arms that are not excluded from being played according to their confidence bounds. Assuming the confidence bounds hold, then if $\mathcal{J}(\mathbf{s}_t)$ contains only one arm then a feasible algorithm incurs no regret in that round. Otherwise, let $i^*$ and $j^*$ be the pair of valid arms with the largest difference in payout for $\mathbf{s}_t$:
$$
i^*,j^* = \argmax_{i,j\in\mathcal{J}(\mathbf{s}_t)} |\mathbf{s}^\mathsf{T}_t(\theta^i-\theta^j)|.
$$
The optimal arm is one of the arms in $\mathcal{J}(\mathbf{s}_t)$, as is the choice of a feasible algorithm, thus this is a bound for the regret in round $t$. Without loss of generality, let $i^*$ be the arm with the higher estimated payoff for context $\mathbf{s}_t$. Since $j^*$ is not excluded, the lower bound of arm $i^*$ is below the upper bound of arm $j^*$. This means that:
\begin{align}
\notag
\mathbf{s}_t\hat{\theta}_t^{i^*} - \sqrt{\beta_t}\|\mathbf{s}_t\|_{(V_t^{i^*})^{-1}} &\le \mathbf{s}_t\hat{\theta}_t^{j^*} + \sqrt{\beta_t}\|\mathbf{s}_t\|_{(V_t^{j^*})^{-1}}\\
\label{eq:confint}\Rightarrow \left|\mathbf{s}_t(\hat{\theta}_t^{i^*} - \hat{\theta}_t^{j^*})\right| &\le \sqrt{\beta_t^{i^*}}\|\mathbf{s}_t\|_{(V_t^{i^*})^{-1}} + \sqrt{\beta_t^{j^*}}\|\mathbf{s}_t\|_{(V_t^{j^*})^{-1}}
\end{align}

Thus:
\begin{align*}
    r_t &\le |\mathbf{s}_t^\mathsf{T}(\theta^{i^*}-\theta^{j^*})|\\
    &\le |\mathbf{s}_t^\mathsf{T}(\theta^{i^*}-\hat{\theta}_t^i)| + |\mathbf{s}_t^\mathsf{T}(\theta^{j^*}-\hat{\theta}_t^{j^*})| + |\mathbf{s}_t^\mathsf{T}(\hat{\theta}_t^{i^*}-\hat{\theta}_t^{j^*})|\\
    &\le |\mathbf{s}_t^\mathsf{T}(\theta^{i^*}-\hat{\theta}_t^{i^*})| + |\mathbf{s}_t^\mathsf{T}(\theta^{j^*}-\hat{\theta}_t^{j^*})| + \sqrt{\beta_t}\|\mathbf{s}_t\|_{(V_t^{i^*})^{-1}} + \sqrt{\beta_t}\|\mathbf{s}_t\|_{(V_t^{j^*})^{-1}}\\
    &\le 
    \|\mathbf{s}_t\|_{(V_t^{i^*})^{-1}}\|(\theta^i-\hat{\theta}_t^{i^*})\|_{V_t^{i^*}} + 
    \|\mathbf{s}_t\|_{(V_t^{j^*})^{-1}}\|(\theta^{j^*}-\hat{\theta}_t^{j^*})\|_{V_t^{j^*}} +
    \sqrt{\beta_t}\|\mathbf{s}_t\|_{(V_t^{i^*})^{-1}} + \sqrt{\beta_t}\|\mathbf{s}_t\|_{(V_t^{j^*})^{-1}}\\
    &\le  2\sqrt{\beta_t}\|\mathbf{s}_t\|_{(V_t^{i^*})^{-1}} + 2\sqrt{\beta_t}\|\mathbf{s}_t\|_{(V_t^{j^*})^{-1}}\\
    &\le 4\sqrt{\beta_t}\|\mathbf{s}_t\|_{(V_t)^{-1}},
    \end{align*}
using: Equation \eqref{eq:confint}; the Cauchy-Schwarz inequality; the fact that $\theta \in \mathcal{C}_t$; and the fact that $\|\mathbf{s}\|^2_{(V_t)^{-1}} = \sum_{i=1}^k\|\mathbf{s}\|^2_{(V_t^i)^{-1}}$, and thus that $\|\mathbf{s}\|_{(V_t^i)^{-1}} \le \|\mathbf{s}\|_{(V_t)^{-1}}$ for any $i$. Up to a factor of 2, this is exactly the form of the regret bound obtained in the proof of Theorem 19.2 \citep{Szepesvari2018}. Thus the rest of the proof proceeds identically from there and establishes the result.
\end{proof}
\end{customthm}

\begin{customthm}{2}
Any rarely-switching algorithm that maintains $(\tilde{\theta}_t^{i}-\tilde{\theta}_t^{j})\in\cone(\Psi_t^{ij})$ for all $i,j\in[k]$ and for all $t$ is a feasible algorithm. 
\begin{proof}

A feasible algorithm never plays an excluded arm. This can be judged on a pairwise basis: if the expected payoff for arm $j$ is lower than that of any other arm $i$, then $j$ is excluded. For a given context $\mathbf{s}_t$, all excluded arms can be determined by considering all pairs of arms. Thus we need only focus on characterizing the decision boundaries between pairs of arms. The confidence sets are constructed such that these decision boundaries lie in \emph{decision regions} with high probability. For two arms, $i$ and $j$, let
$\Psi^{ij}_t$ be the set difference $\Psi^{ij}_t = \{\psi\in\mathbb{R}^d| \psi = \theta^i - \theta^j, \text{with } \theta \in \mathcal{C}_t\}$. 

Then the decision boundary between arms $i$ and $j$ lies in the set:
$$
\mathcal{D}^{ij}_t = \{\mathbf{s}\in\mathbb{R}^d| \mathbf{s}^\mathsf{T}\psi^{ij} = 0, \psi^{ij} \in \Psi^{ij}_t\}.
$$
This region can be understand as the complement of the union of two sets: the dual cone of $\Psi^{ij}_t$ and the polar cone of $\Psi^{ij}_t$. The dual cone corresponds to contexts in which, for any set of feasible parameters $\theta \in\mathcal{C}_t$, arm $i$ is better: $C^{*ij}_t \coloneqq \{\mathbf{s}\in\mathbb{R}^d|\mathbf{s}^\mathsf{T}\psi^{ij} > 0, \forall \psi^{ij}\in\Psi^{ij}_t\}$. The polar cone, $D_t^{*ij} \coloneqq -C_t^{*ij}$ is the opposite: it represents contexts for which, given any combination of feasible parameters, arm $j$ is better. 

If $\tilde{\theta}^i-\tilde{\theta}^j \in \cone(\Psi_t^{ij})$ then $\exists \alpha > 0$ such that 
$$
\tilde{\theta}^i - \tilde{\theta}^j = \alpha(\theta^i-\theta^j)
$$
for some $\theta^i,\theta^j\in\mathcal{C}_t$. This means there exists feasible parameters that have the same decision boundary as $\tilde{\theta}^i,\tilde{\theta}^j$. This means that, for all $\mathbf{s}\in C_t^{*ij}$, where arm $\theta^i$ is better, the policy $\tilde{\theta}$ will not play arm $j$ (the excluded arm). Thus $\tilde{\theta}$ does not play any excluded arm, and is a feasible algorithm. Note that if $0\in \Psi_t^{ij}$ then any decision boundary between $i$ and $j$ is plausible, and this pair of arms does not constrain if $\tilde{\theta}$ is feasible or not.


\end{proof}
\end{customthm}

Finally, we can prove in $d = 2$ there are updates that always improve the expected regret. The intuition is that, in two dimensions, the policy $\tilde{\theta}_t$ defines intervals over angles in which each arm is played. A policy thus becomes one dimensional -- it can be considered as a cyclic ordering of the arms based on angle, $\sigma = (\sigma(\tilde{\theta}_1)\dots\sigma(\tilde{\theta}_k))$, and a set of angles corresponding to the decision boundaries between the pair of arms $\tilde{\theta}^{\sigma_{i}}$ and $\tilde{\theta}^{\sigma_{i+1}}$.

\begin{figure*}
\centering
\includegraphics[width=0.6\textwidth]{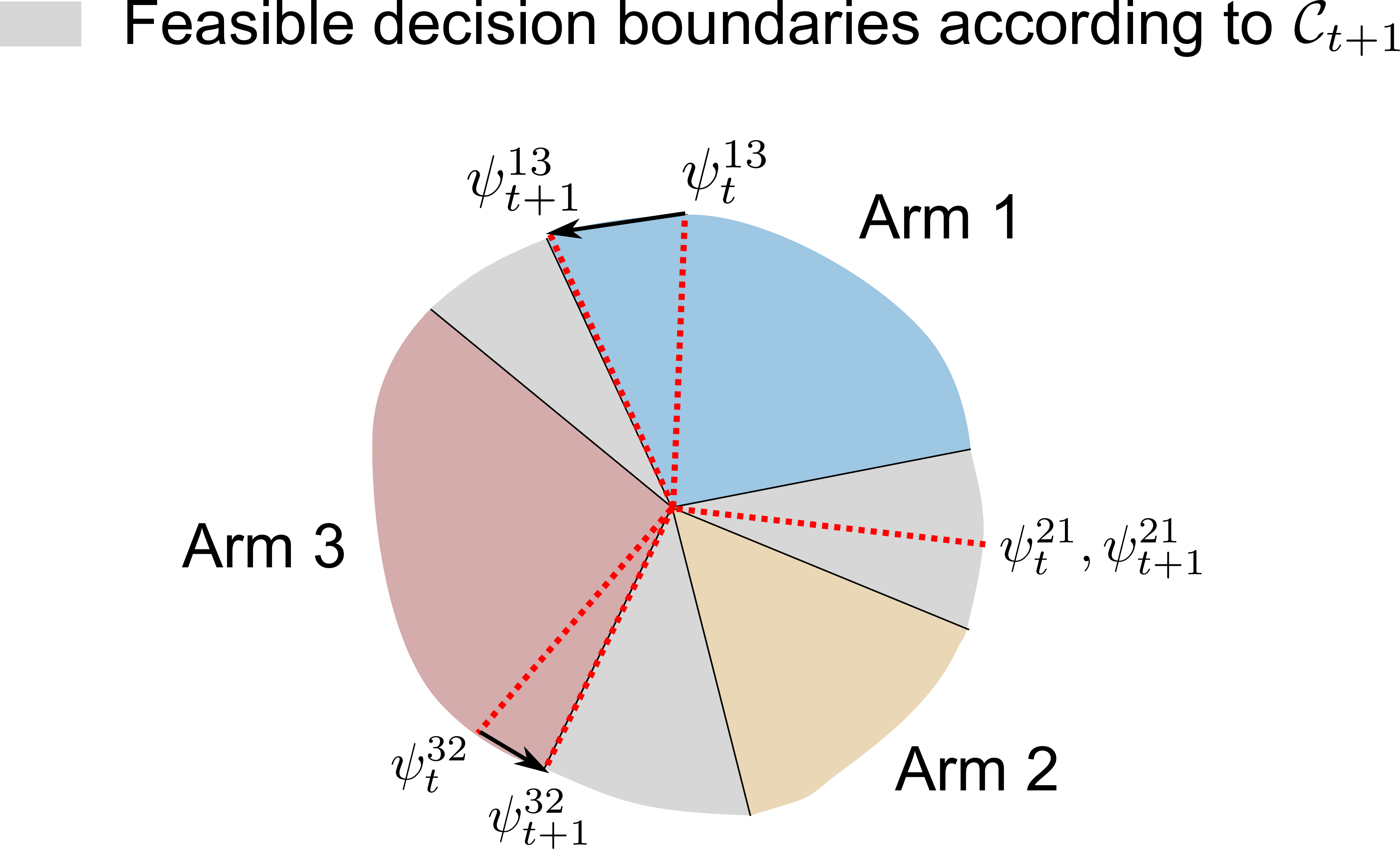}
\caption{Example of changing policy from $\tilde{\theta}_t$ to $\tilde{\theta}_{t+1}$. The decision boundaries for both $\tilde{\theta}_t$ and $\tilde{\theta}_{t+1}$ are shown in dashed red. Combined they form the set $\Omega$. Assuming the conditions in Theorem 3, the regions where the policy changes its behavior are regions where regret was definitely being incurred (assuming the confidence bounds are true) -- indicated by the black arrows. Thus these updates can not increase the expected regret.}
\label{fig:angles}
\end{figure*}

For a given change in policy $\tilde{\theta}_t\to\tilde{\theta}_{t+1}$ consider the set, $\Omega$, of angles that describe decision boundaries in both the old and the new policies (Supplementary Figure \ref{fig:angles}). The set of angles $\Omega$ gives a cyclic ordering to the combined new and old decision boundaries. Then we have:
\begin{customthm}{3}
For $d = 2$, if an update to the policy $\tilde{\theta}_t$ satisfies:
\begin{enumerate}[label=(\roman*)]
    \item the set of playable arms stays the same between $t$ and $t+1$ -- that is, an arm does not become dominated (stops being an extreme point of the polytope), or stops being dominated;
    \item the ordering of arms, $\sigma$, in the new and old policies is the same;
    \item for each decision boundary $b = \psi_t^{\sigma_i,\sigma_{i+1}}$, either the element before or after $b$ in $\Omega$ is $\psi_{t+1}^{\sigma_i,\sigma_{i+1}}$;
    \item each change $\psi_{t}^{ij}\to\psi_{t+1}^{ij}$ is the smallest such that $\psi_{t+1}^{ij}\in\cone(\Psi_{t+1}^{ij})$. I.e. that
    $$
    \psi_{t+1}^{ij} = \argmin_{\psi\in\cone(\Psi_{t+1}^{ij})}\text{arccosdist}(\psi,\psi_t^{ij});
    $$
\end{enumerate}
then it does not increase the expected instantaneous regret, for a fixed policy, in the sense that:
$$
\mathbb{E}_{\mathbf{s}\sim \rho}\left(\mathbf{s}^\mathsf{T} \theta^{a_t}\right)
\le \mathbb{E}_{\mathbf{s}\sim \rho}\left(\mathbf{s}^\mathsf{T} \theta^{a_{t+1}}\right),
$$
with probability at least $1-\delta$.

\begin{proof}
In the planar case, edges constitute the faces of the polytope, thus specifying the decision boundaries (edges) is the same as the H-representation of the polytope -- it uniquely specifies the polytope. Thus we can equally think of dealing with vertices (arm parameters), or decision boundaries. Here we deal entirely with the decision boundaries.

Note first that if $0\in\Psi_{t+1}^{ij}$ then any boundary $\psi_{t+1}^{ij}$ is feasible, and this does not contribute to changing the policy.

As usual, throughout the proof, assume that the confidence bounds hold (true with probability at least $1-\delta$).

We prove the result by considering each interval $[\psi_{t}^{\sigma_i,\sigma_{i+1}},\psi_{t+1}^{\sigma_i,\sigma_{i+1}}]$. These intervals describe contexts where the policy changes arms. By the assumptions of the update, no such interval overlaps with any other interval. Since the behavior of $\tilde{\theta}_t$ and $\tilde{\theta}_{t+1}$ is the same outside of these intervals then to establish that the change in policy does not incur more regret we just need to study the behavior of the policy in each of these intervals separately. 

So, for contexts in a given interval, between $t$ and $t+1$, the policy switches from playing $\sigma_{i+1}$ to $\sigma_{i}$ (or vice versa, depending on the ordering). Assume the policy switches from $\sigma_{i+1}$ to $\sigma_{i}$; the same argument applies to the opposite case. Either the interval is empty, and the policy behaves the same between $t$ and $t+1$, or by condition (iv) the decision boundary is moved just to the edge of the feasible boundary between arms $\sigma_i$ and $\sigma_{i+1}$.

Assume the policy is moved just to the edge of the feasible boundary between the arms. The movement to the boundary of the feasible region means that the interval cannot contain the true decision boundary $\sigma_{i},\sigma_{i+1}$. Thus, by the conditions, in the interval there is no way that playing $\sigma_{i+1}$ was feasible. The old policy played $\sigma_{i+1}$ in this interval, and so only incurred regret. By changing the policy to play $\sigma_i$ in this interval, expected regret cannot increase.
\end{proof}
\end{customthm}

\textbf{Remark:} The assumptions are not likely to be true at the start of the trial, where the ordering of the arms is unclear and the uncertain policy regions are not isolated from one another. However these assumptions may be reasonable later, in which case they provide a guarantee that the bandit never decreases its performance. By design, a conservative rarely-switching algorithm can be implemented to satisfy condition (iv) for all policy updates. Thus it is just the remaining conditions that need to be checked.

\section{Comparison to a deterministic bandit algorithm}

\begin{figure*}
\centering
\includegraphics[width=0.8\textwidth]{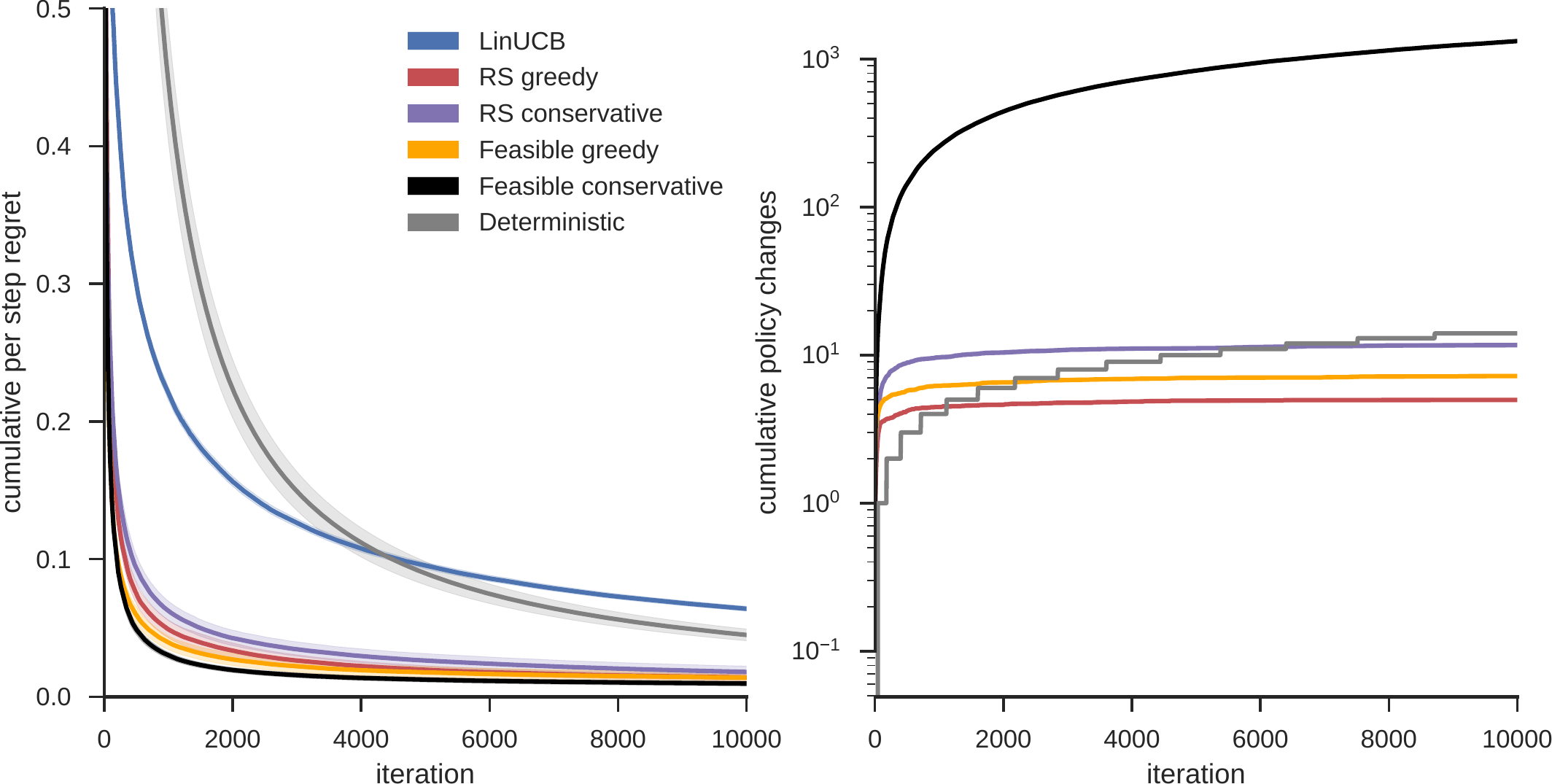}
\caption{Deterministic bandit algorithm performance. Policy is updated deterministically (grey curve on right panel). Regret is higher for a deterministic policy (grey curve on left panel), despite it making more policy changes by the end of the trial than the other rarely-switching algorithms. For clarity not all other bandit algorithms are plotted here. Traces show mean plus/minus standard error.}
\label{fig:det_bandit}
\end{figure*}

We can compare the rarely-switching policies to a deterministic alternative. The deterministic algorithm updates policy on a fixed schedule such that the total number of policy changes as a function of round number $t$ is proportional to $\sqrt{t}$. This is chosen to roughly match the behavior of the rarely-switching algorithms. At each policy change, the parameters are updated to $\tilde{\theta}_t = \hat{\theta}_t$. It is clear that, even though by the end of the trial the deterministic algorithm has made more policy changes than the rarely-switching algorithms, the cumulative regret is higher (Supplementary Figure \ref{fig:det_bandit}). These results suggest it is worthwhile considering an adaptive rarely-switching algorithm, that decides when to update its policy based on current confidence estimates of each arm.

\section{Extra simulation results}

\begin{figure*}
\centering
\includegraphics[width=\textwidth]{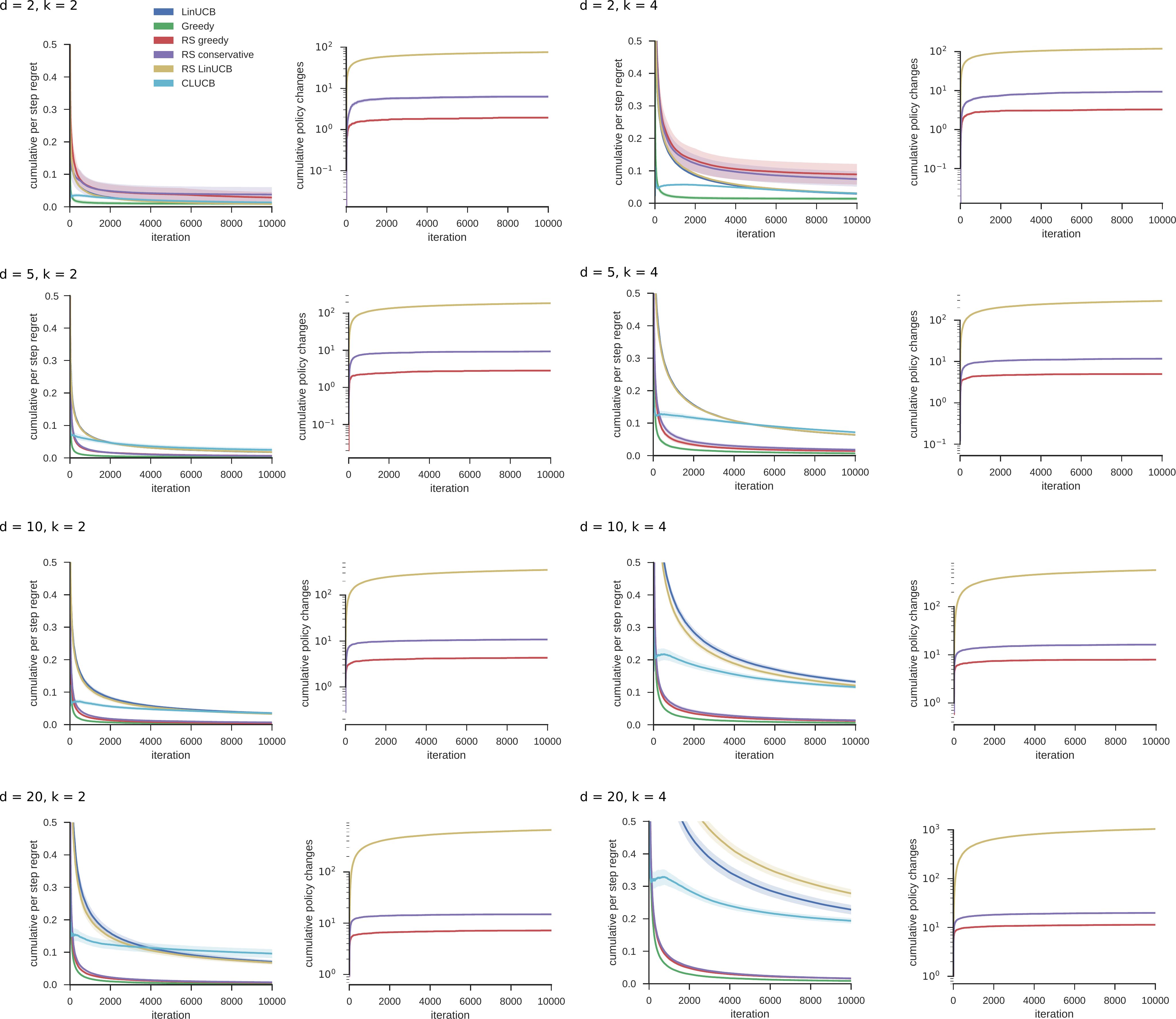}
\caption{Simulation results varying dimension $d$ and number of arms $k$. Fifty random generated bandit parameters are run for 10000 rounds. Plotted are the per-step regret ($R_t/t$) and the number of policy changes. Traces show mean, plus/minus standard error.}
\label{fig:simulated_comparisons}
\end{figure*}

We perform the same simulations as in the main text, except vary the number of dimensions $d$ and the number of arms $k$. We find similar results for $d \ge 5$. For $d = 2$, the rarely switching bandits do not perform as well as LinUCB (Supplementary Figure \ref{fig:simulated_comparisons}). 

\end{document}